\pdfoutput=1
\documentclass[11pt]{article}
\usepackage{acl}
\usepackage{placeins}
\usepackage{float}
\usepackage{times}
\usepackage{latexsym}
\usepackage{booktabs}
\usepackage[T1]{fontenc}
\usepackage[utf8]{inputenc}
\usepackage{microtype}
\usepackage{amsmath}
\usepackage{inconsolata}
\usepackage{algorithm}
\usepackage{algorithmicx}
\usepackage{algpseudocode}
\usepackage{graphicx}

\title{MSA at SemEval-2025 Task 3: High Quality Weak Labeling and LLM Ensemble Verification for Multilingual Hallucination Detection}

\author{Baraa Hikal, Ahmed Nasreldin, Ali Hamdi \\
  Faculty of Computer Science, MSA University, Egypt \\
  \texttt{\{baraa.moaweya, ahmed.nasreldin, ahamdi\}@msa.edu.eg}
}

\begin{document}
\maketitle
\footnotetext[1]{\url{https://github.com/baraahekal/mu-shroom}}

\begin{abstract}
This paper describes our submission for SemEval-2025 Task 3: Mu-SHROOM, the Multilingual Shared-task on Hallucinations and Related Observable Overgeneration Mistakes \cite{vazquez-etal-2025-mu-shroom}. The task involves detecting hallucinated spans in text generated by instruction-tuned Large Language Models (LLMs) across multiple languages. Our approach combines task-specific prompt engineering with an LLM ensemble verification mechanism, where a primary model extracts hallucination spans and three independent LLMs adjudicate their validity through probability-based voting. This framework simulates the human annotation workflow used in the shared task validation and test data. Additionally, fuzzy matching refines span alignment. Our system ranked 1\textsuperscript{st} in Arabic and Basque, 
2\textsuperscript{nd} in German, Swedish, and Finnish, 
and 3\textsuperscript{rd} in Czech, Farsi, and French.
\end{abstract}

\section{Introduction}

Large Language Models (LLMs) are highly effective in generating text; however, they sometimes produce hallucinations—misleading content that is not properly grounded in the input data \cite{huang2025survey}. Identifying these spans is essential for improving the reliability of LLM-generated outputs in translation, summarization, and conversational AI \cite{alaharju2024ensuring}. SemEval-2025 Task 3: Mu-SHROOM tackles this challenge by presenting a multilingual benchmark for detecting character-level hallucinations across multiple languages. The task involves detecting hallucinated spans in instruction-tuned LLM outputs, presenting challenges in language diversity, annotation consistency, and accurate span localization. \cite{sriramanan2025llm}

To tackle this challenge, our system utilizes a hybrid approach that integrates task-specific prompt engineering for weak label generation with an LLM ensemble verification mechanism \cite{hikal2025few}. Our methodology follows a multi-step adjudication process in which a primary LLM identifies hallucination spans, and three independent LLMs subsequently verify their validity through a probability-based voting mechanism \cite{kang2024using}. Additionally, we apply fuzzy matching techniques to improve the alignment of hallucination spans with ground truth annotations, thereby enhancing detection accuracy \cite{chaudhuri2003robust}.

By participating in this task, we gained insights into language-specific hallucination challenges and the strengths and limitations of LLM-based verification. Certain LLMs demonstrated closer alignment with human annotations, while hallucination patterns varied significantly, particularly in morphologically rich languages where annotation ambiguity was higher \cite{abdelrahman2024hallucination}. Our results indicate that ensemble verification and span refinement substantially improve hallucination detection, offering a robust approach for mitigating LLM hallucinations in multilingual settings.

\section{Related Work}

Hallucination detection in Large Language Models (LLMs) has been studied in machine translation, text summarization, and conversational AI \cite{ji2023survey}. Earlier approaches primarily relied on sentence-level classification, whereas recent research has transitioned to span-level detection for greater precision \cite{joshi2020spanbert}. Self-consistency verification and knowledge-grounded approaches have improved hallucination identification, but many depend on external data, limiting their applicability in multilingual settings. \cite{mehta2024halu}

Multilingual NLP models struggle with hallucinations, especially in low-resource languages where confidence scores are unreliable \cite{kang2024comparing}. Morphologically rich languages introduce additional challenges due to intricate annotation inconsistencies \cite{tsarfaty2013parsing}. Prior work on translation-based verification has attempted to address this, but these approaches are ineffective in zero-shot scenarios \cite{nie2022zero}.

Ensemble verification methods enhance detection accuracy by utilizing multiple models. Approaches such as multi-agent verification and cross-model adjudication have proven effective in assessing LLM outputs \cite{liu2024genotex}. Our system expands on these approaches by integrating weak label generation with an ensemble verification pipeline, while also utilizing fuzzy matching to improve span alignment. Unlike previous methods that rely on single-model hallucination detection, our approach leverages an ensemble of LLMs for adjudication, reducing model bias and improving hallucination span refinement via fuzzy matching.

\section{System Overview}

\begin{figure*}[ht]
    \centering
    \includegraphics[width=0.85\linewidth]{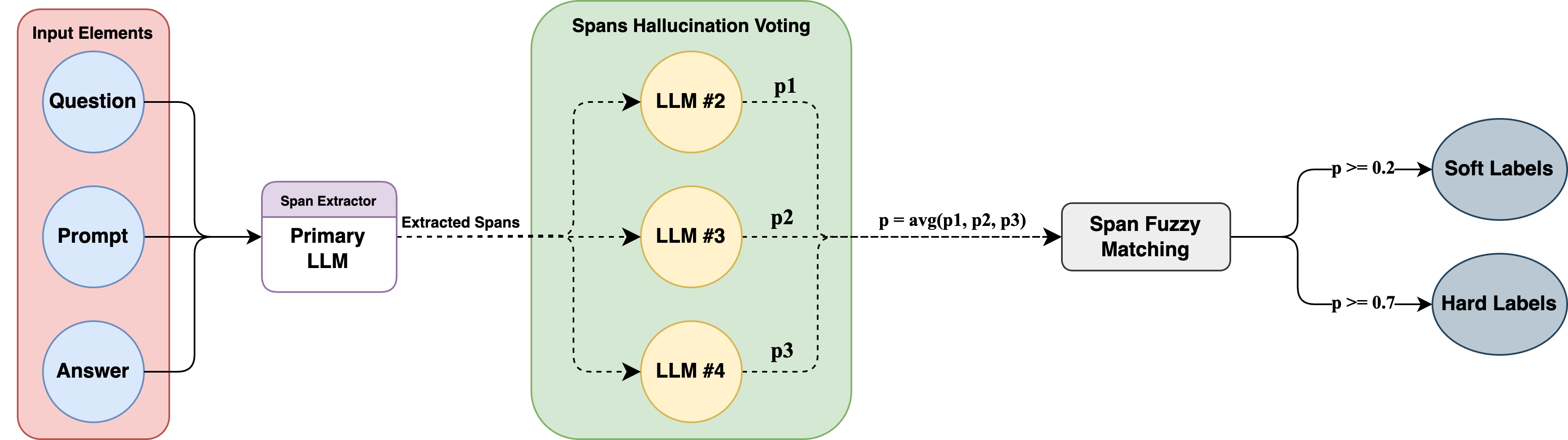}
    \caption{Overview of our hallucination detection pipeline.}
    \label{fig:pipeline_diagram}
\end{figure*}

Our hallucination detection approach integrates task-specific prompt engineering, an LLM ensemble verification mechanism, and post-processing refinements. The system is composed of three key components: fine-tuned prompt construction, hallucination span verification through LLM ensembles, and post-processing with fuzzy matching. An overview of the full pipeline is illustrated in Figure~\ref{fig:pipeline_diagram}.

\subsection{Prompt Engineering for Weak Label Generation}

We analyzed the validation dataset to extract annotator instructions and identify patterns, enabling the construction of a fine-grained prompt with few-shot examples. Iterative refinement improved extraction accuracy. Detailed prompt in Appendix~\ref{appendix:annotation}.

\subsection{Selection of State-of-the-Art LLMs}

\begin{figure*}[ht]
    \centering
    \includegraphics[width=0.8\linewidth]{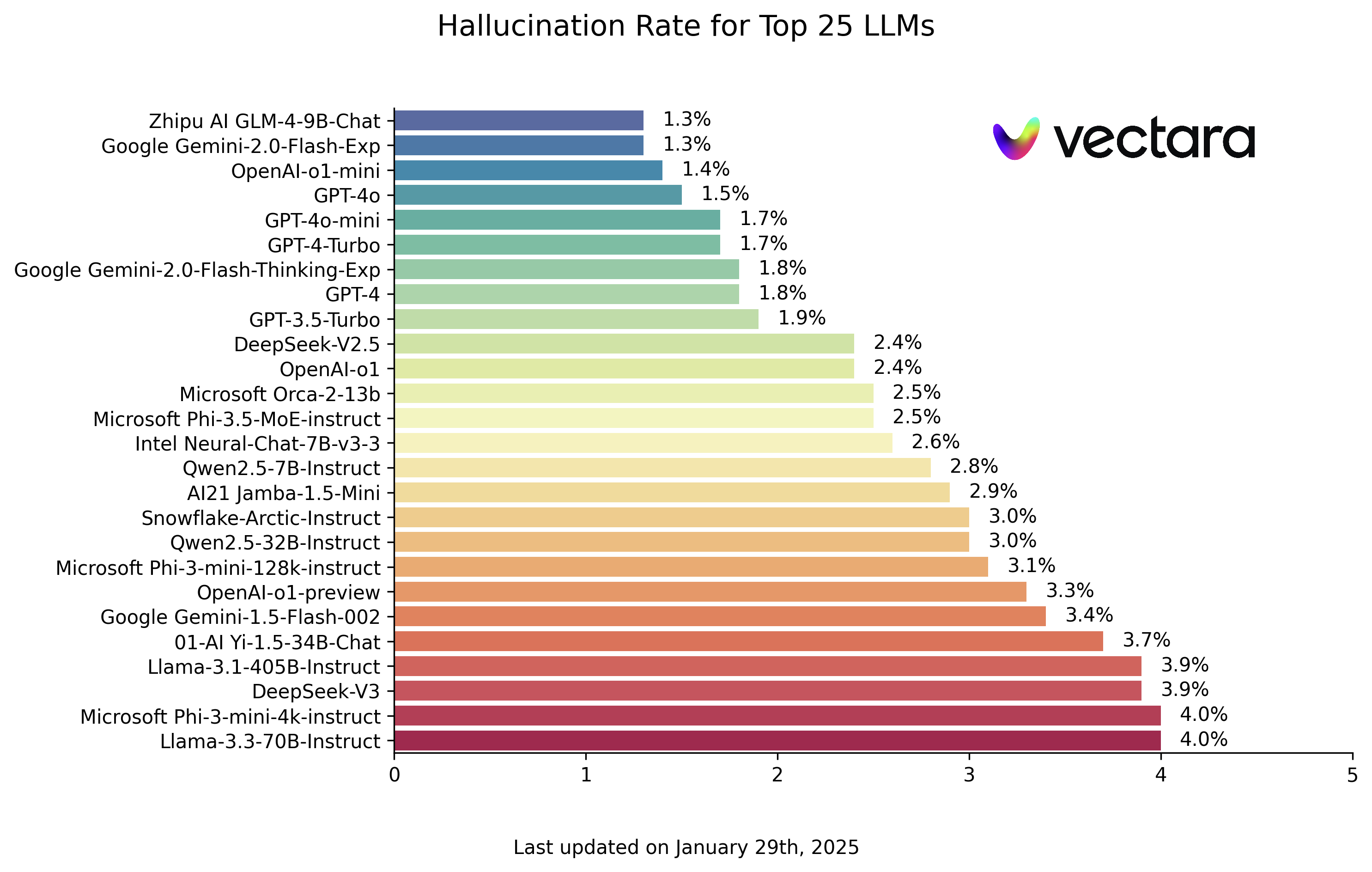}
    \caption{Performance rankings of LLMs according to the Vectara Hallucination Leaderboard~\cite{vectara2024hallucination}.}
    \label{fig:vectara_report}
\end{figure*}

Building on the insights from the Vectara LLM Report, we chose Gemini-2.0-Flash-Exp, Qwen-2.5-Max~\cite{qwen2024technical}, GPT-4o~\cite{openai2024gpt4o}, and DeepSeek-V3~\cite{deepseek2024v3} as our primary models for hallucination detection. These models were selected for their strong factual accuracy and reliable generation capabilities, ensuring consistent performance across multiple languages. Figure~\ref{fig:vectara_report} illustrates the model rankings from the report.

\subsection{LLM Ensemble Verification Mechanism}
Our hallucination detection pipeline utilizes a multi-stage ensemble verification process. With four selected LLMs—Gemini-2.0-Flash-Exp, Qwen-2.5-Max, GPT-4o, and DeepSeek-V3—we systematically rotate through different configurations, where one model identifies hallucinated spans while the other three act as adjudicators.
This setup is inspired by the Mu-SHROOM annotation process, where multiple human annotators reviewed and adjudicated hallucination spans in the validation and test datasets. By simulating this human adjudication process with LLMs, we aim to improve label consistency and mitigate annotation biases.

\paragraph{Span Extractor Model (SEM)} A primary LLM identifies hallucinated spans by analyzing question-answer pairs. Given a question \( Q \) and an answer \( A \), the span extractor outputs candidate hallucination spans $S = \{s_1, s_2, \dots, s_k\}$:

\[
S = \text{LLM}_{\text{extract}}(Q, A, \text{prompt})
\]

\paragraph{Voting Adjudicator Models (VAMs)} The three remaining LLMs act as adjudicators, independently assessing each span $s_i \in S$ and assigning a hallucination probability score:

\[
p_{ij} = M_j(s_i, Q), \quad p_{ij} \in [0,1]
\]

where $M_j$ represents an adjudicator LLM.

\paragraph{Iterative Model Rotation:} This process is repeated for all possible combinations of the four models, ensuring that each model serves as the span extractor exactly once, while the other three act as adjudicators. Given four models, this results in a total of four unique verification runs.

\paragraph{Consensus-Based Labeling (CBL):} The final hallucination probability for each span is determined by aggregating the probabilities across all verification runs:
\[
p_i = \frac{1}{N} \sum_{j=1}^{N} p_{ij}
\]
where $N=3$ is the number of adjudicator models per run. The final hallucination label is assigned using a majority voting scheme across all runs. A span is classified as hallucinated if:
\[
p_i \geq 0.7
\]
The threshold $\tau = 0.7$ was chosen based on empirical observations on the validation set. During tuning, we found that lower thresholds (e.g., 0.5) tended to produce too many false positives by labeling uncertain spans as hallucinations, while higher thresholds (e.g., 0.8) missed subtle hallucinations annotated by human reviewers. A threshold of 0.7 offered the best trade-off between precision and recall, and its behavior closely matched the annotation patterns observed in the Mu-SHROOM validation data~\cite{vazquez-etal-2025-mu-shroom}.

This iterative model selection ensures robustness by reducing individual model biases and leveraging diverse perspectives from different LLMs.

\subsection{Post-Processing with Fuzzy Matching}

LLMs frequently introduce minor inconsistencies in span extraction, such as variations in capitalization, extra spaces, or incomplete word boundaries. To minimize these errors, we use fuzzy matching with a similarity threshold of 0.9 (partial ratio). The similarity score between a predicted span $s_i$ and a ground truth span $g_j$ is given by:
\[
\text{Similarity}(s_i, g_j) = 1 - \frac{\text{Lev}(s_i, g_j)}{\max(|s_i|, |g_j|)}
\]
where $\text{Lev}(s_i, g_j)$ is the Levenshtein distance. If $\text{Similarity}(s_i, g_j) \geq 0.9$, the span is considered correctly aligned.

\subsection{Algorithm Implementation}

Our pipeline follows a multi-stage verification process where a primary LLM extracts candidate hallucination spans, and three adjudicator models verify them using probability-based voting. Fuzzy matching refines span alignment, improving precision. This ensemble approach mitigates model bias and enhances robustness.

Algorithm~\ref{alg:hallucination_detection} in Appendix~\ref{appendix:algorithm} outlines the full process.

\begin{figure*}[ht]
    \centering
    \includegraphics[width=\textwidth]{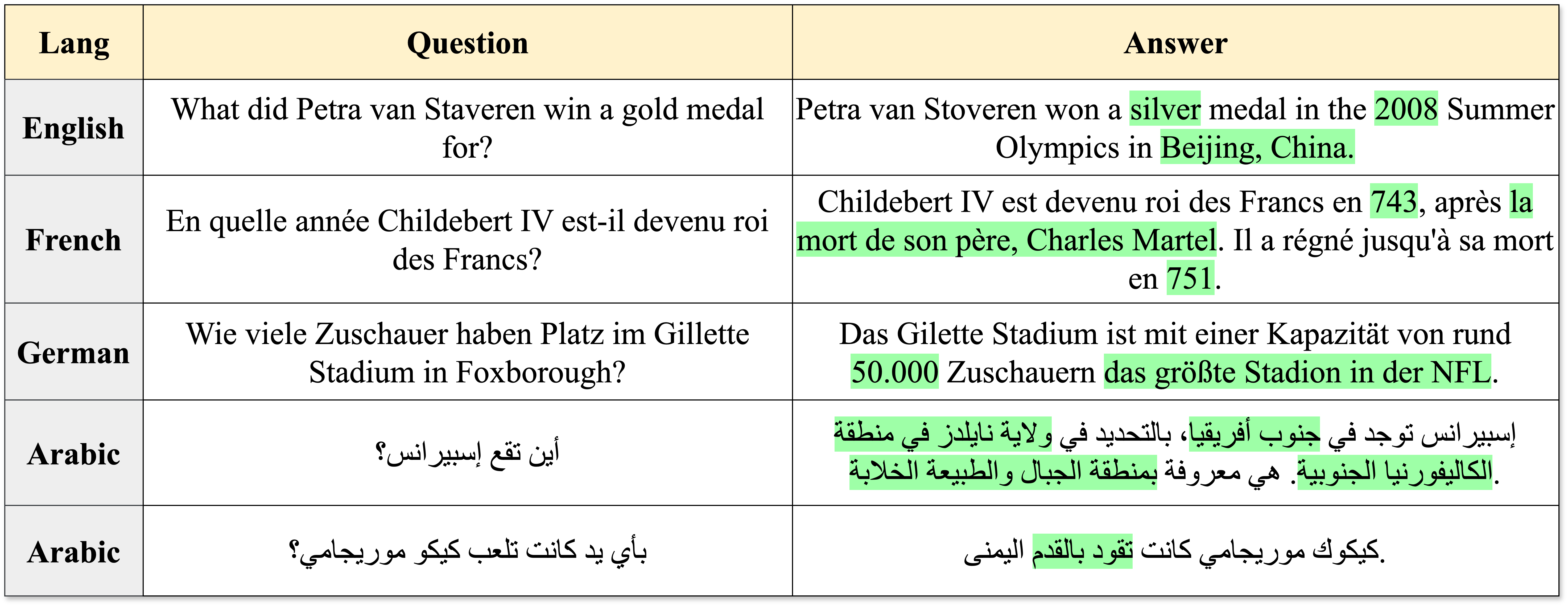}
    \caption{Dataset examples in different languages. The hallucinated span(s) are highlighted.}
    \label{fig:dataset_example}
\end{figure*}

\section{Experiments and Results}

\subsection{Dataset}
Our system was evaluated on the Mu-SHROOM dataset from SemEval-2025 Task 3. We leveraged only the validation and test sets, using the validation set for prompt refinement and the test set for final evaluation. Unlike traditional supervised approaches, we did not use the training set for model learning. Instead, we employed prompt-based weak labeling and an ensemble verification mechanism \cite{smith2024language}. The test set contained unlabeled examples, and final system evaluation was conducted by the task organizers.

Figure~\ref{fig:dataset_example} presents dataset examples in different languages, highlighting hallucinated spans.

\subsection{Evaluation Metrics}
We evaluated our system using the official Mu-SHROOM metrics:
\begin{itemize}
    \item \textbf{Intersection-over-Union (IoU)}: Measures the overlap between predicted and gold hallucinated spans \cite{rezatofighi2019generalized}.
    \item \textbf{Probability Correlation (Corr)}: Evaluates the correlation between predicted hallucination probabilities and human annotations \cite{sheugh2015note}.
\end{itemize}

The IoU score for a predicted span $s_p$ and a ground truth span $s_g$ is computed as:
\[
\text{IoU} = \frac{|s_p \cap s_g|}{|s_p \cup s_g|}
\]
where $|s_p \cap s_g|$ represents the overlapping characters, and $|s_p \cup s_g|$ is the total number of unique characters in both spans.

\subsection{Results}
As each of the four LLMs alternates as the span extractor while the others act as adjudicators, we report results for each combination. The tables [\ref{tab:qwen_results},\ref{tab:gemini_results},\ref{tab:gpt4o_results},\ref{tab:deepseek_results}] show performance across languages.

\FloatBarrier
\begin{table}[ht]
    \centering
    \small
    \begin{tabular}{l|c|c}
        \toprule
        \textbf{Lang} & \textbf{IoU Score} & \textbf{Probability Corr} \\
        \midrule
        AR  & 0.576 & 0.536 \\
        EU  & 0.604 & 0.611 \\
        DE  & 0.526 & 0.567 \\
        SV  & 0.607 & 0.401 \\
        FI  & 0.587 & 0.501 \\
        CS  & 0.396 & 0.410 \\
        FA  & 0.540 & 0.511 \\
        FR  & 0.571 & 0.507 \\
        EN  & 0.506 & 0.538 \\
        IT  & 0.484 & 0.545 \\
        HI  & \textbf{0.684} & 0.725 \\
        \bottomrule
    \end{tabular}
    \caption{Performance when Qwen-2.5-Max acts as the span extractor.}
    \label{tab:qwen_results}
\end{table}
\FloatBarrier

\FloatBarrier
\begin{table}[ht]
    \centering
    \small
    \begin{tabular}{l|c|c}
        \toprule
        \textbf{Lang} & \textbf{IoU Score} & \textbf{Probability Corr} \\
        \midrule
        AR  & \textbf{0.669} & 0.648 \\
        EU  & \textbf{0.612} & 0.620 \\
        DE  & 0.601 & 0.547 \\
        SV  & \textbf{0.636} & 0.422 \\
        FI  & 0.625 & 0.521 \\
        CS  & \textbf{0.507} & 0.552 \\
        FA  & \textbf{0.669} & 0.679 \\
        FR  & \textbf{0.619} & 0.555 \\
        EN  & \textbf{0.531} & 0.519 \\
        IT  & 0.712 & 0.737 \\
        HI  & 0.662 & 0.690 \\
        \bottomrule
    \end{tabular}
    \caption{Performance when Gemini-2.0-Flash-Exp acts as the span extractor.}
    \label{tab:gemini_results}
\end{table}
\FloatBarrier

\FloatBarrier
\begin{table}[ht]
    \centering
    \small
    \begin{tabular}{l|c|c}
        \toprule
        \textbf{Lang} & \textbf{IoU Score} & \textbf{Probability Corr} \\
        \midrule
        AR  & 0.637 & 0.593 \\
        EU  & 0.604 & 0.611 \\
        DE  & 0.527 & 0.531 \\
        SV  & 0.610 & 0.398 \\
        FI  & 0.619 & 0.527 \\
        CS  & 0.432 & 0.486 \\
        FA  & 0.639 & 0.700 \\
        FR  & 0.601 & 0.485 \\
        EN  & 0.525 & 0.502 \\
        IT  & \textbf{0.736} & 0.756 \\
        HI  & 0.621 & 0.664 \\
        \bottomrule
    \end{tabular}
    \caption{Performance when GPT-4o acts as the span extractor.}
    \label{tab:gpt4o_results}
\end{table}
\FloatBarrier

\FloatBarrier
\begin{table}[ht]
    \centering
    \small
    \begin{tabular}{l|c|c}
        \toprule
        \textbf{Lang} & \textbf{IoU Score} & \textbf{Probability Corr} \\
        \midrule
        AR  & 0.658 & 0.644 \\
        EU  & 0.607 & 0.585 \\
        DE  & \textbf{0.613} & 0.610 \\
        SV  & 0.624 & 0.417 \\
        FI  & \textbf{0.642} & 0.546 \\
        CS  & 0.465 & 0.507 \\
        FA  & 0.632 & 0.671 \\
        FR  & 0.572 & 0.539 \\
        EN  & 0.529 & 0.487 \\
        IT  & 0.703 & 0.716 \\
        HI  & 0.659 & 0.697 \\
        \bottomrule
    \end{tabular}
    \caption{Performance when DeepSeek-V3 acts as the span extractor.}
    \label{tab:deepseek_results}
\end{table}
\FloatBarrier

\FloatBarrier
\begin{table}[ht]
    \centering
    \small
    \setlength{\tabcolsep}{4pt} % Adjust column spacing for a tighter fit
    \begin{tabular}{l|c|c|c|c}
        \toprule
        \textbf{Lang} & \textbf{Span Extractor} & \textbf{IoU} & \textbf{Corr} & \textbf{Rank} \\
        \midrule
        AR  & Gemini-2.0-Flash-Exp  & 0.669  & 0.648  & 1/32 \\
        EU  & Gemini-2.0-Flash-Exp  & 0.612  & 0.620  & 1/26 \\
        DE  & DeepSeek-V3  & 0.613  & 0.610  & 2/31 \\
        SV  & Gemini-2.0-Flash-Exp  & 0.636  & 0.422  & 2/30 \\
        FI  & DeepSeek-V3  & 0.642  & 0.546  & 2/30 \\
        CS  & Gemini-2.0-Flash-Exp  & 0.507  & 0.552  & 3/26 \\
        FA  & Gemini-2.0-Flash-Exp  & 0.669  & 0.679  & 3/26 \\
        FR  & Gemini-2.0-Flash-Exp  & 0.619  & 0.555  & 3/33 \\
        IT  & GPT-4o  & 0.736  & 0.756  & 4/31 \\
        HI  & Qwen-2.5-Max  & 0.684  & 0.725  & 5/27 \\
        EN  & Gemini-2.0-Flash-Exp  & 0.531  & 0.519  & 6/44 \\
        \bottomrule
    \end{tabular}
    \caption{Best performance per language, with span extractor and final rank.}
    \label{tab:final_results}
\end{table}
\FloatBarrier
Our system outperformed other methods in Arabic and Basque, where annotation consistency was higher. However, performance dropped in English, likely due to increased annotation variability—English had up to 12 different annotators per sample \cite{vazquez-etal-2025-mu-shroom} leading to inconsistencies.

\subsection{Discussion}
Our system effectively detects hallucinated spans across multiple languages by using ensemble verification to reduce model bias and fuzzy matching to refine span alignment. However, challenges remain—especially in dealing with annotation inconsistencies and ambiguous hallucinations, which tend to be more common in morphologically complex languages.

A key finding is that different LLMs vary in their alignment with human annotations, indicating that task-specific fine-tuning or alternative verification strategies could further improve detection accuracy. Additionally, improving span refinement techniques beyond fuzzy matching may reduce boundary mismatches and improve character-level precision.

\section{Conclusion}
We presented our system for \textit{SemEval-2025 Task 3: Mu-SHROOM}, focusing on hallucinated span detection in LLM-generated text across multiple languages. Our approach combines prompt-engineered weak label generation with an LLM ensemble verification mechanism, demonstrating strong performance in multilingual hallucination detection.

Our results confirm  the effectiveness of ensemble-based adjudication, ranking among the top systems in several languages. However, challenges such as annotation variability and morphological complexity highlight areas for further refinement.

Future work  could focus on integrating external knowledge for hallucination verification, fine-tuning LLMs to better align with human annotations, and refining span localization techniques to enhance character-level precision. These improvements could further advance hallucination detection in multilingual NLP systems.

\vspace{10em}

\appendix
\section{Appendix: Instruction Prompt template for Extraction and Annotation}
\label{appendix:annotation}

\vspace{0.5em}
\hrule
\vspace{0.3em}
\textbf{Question \& Answer Pair}  
\vspace{0.3em}
\hrule
\vspace{0.3em}

\noindent\textbf{i) Question:}  
\texttt{model-input}  
\vspace{0.3em}

\noindent\textbf{ii) Answer:}  
\texttt{model-output-text}  
\vspace{0.5em}

\hrule
\vspace{0.3em}
\noindent\textbf{Task Description}
\vspace{0.3em}
\hrule
\vspace{0.3em}

\noindent\texttt{You are a professional annotator and \{entry[lang]\} linguistic expert. Your job is to detect and extract hallucination spans from the provided answer compared to the question.}  
\vspace{0.5em}

\hrule

\vspace{0.3em}
\noindent\textbf{Exact Span Matching}
\vspace{0.3em}
\hrule
\vspace{0.5em}

\noindent\texttt{Extract spans word-for-word and character-for-character exactly as they appear in the answer. Ensure perfect alignment, including punctuation, capitalization, and spacing. If a span is partially supported, only extract the unsupported portion. Preserve original numeral formats: Persian/Arabic numerals must remain in their native script.}  

\hrule
\vspace{0.3em}
\noindent\textbf{Minimal Spans}
\vspace{0.3em}
\hrule

\vspace{0.5em}
\noindent\texttt{Select the smallest possible spans that, when removed, completely eliminate the hallucination. Prioritize precision: Avoid extracting entire sentences if a shorter phrase accurately captures the hallucination. Ensure the extracted span exclusively contains hallucinated content without removing valid information.}  
\vspace{0.3em}

\hrule
\vspace{0.3em}
\noindent\textbf{Hallucination Definition}
\vspace{0.3em}
\hrule

\vspace{0.5em}
\noindent\texttt{Any phrase, entity, number, or fact that is not supported by the question. Any exaggeration or overly specific detail absent in the question. Incorrect names, locations, numbers, dates, or causes. In yes/no questions, unsupported answers (e.g., "Yes", "No") and speculative details.}  
\vspace{0.5em}

\hrule
\vspace{0.3em}
\noindent\textbf{Soft and Hard Labels}
\vspace{0.3em}
\hrule

\vspace{0.5em}
\noindent\texttt{Assign probabilities [0.0 - 1.0] for soft labels based on hallucination confidence. Include spans with $\geq 0.7$ probability in hard labels.}

\vspace{3em}

\section{Appendix: Our Proposed Framework}
\label{appendix:algorithm}
\begin{algorithm}
\caption{Hallucination Detection Pipeline}
\label{alg:hallucination_detection}
\begin{algorithmic}[1]

\Require Question $Q$, Answer $A$, LLM ensemble $\{M_1, M_2, M_3\}$, threshold $\tau = 0.7$
\Ensure Set of hallucinated spans $S^*$

\State $S \gets \text{LLM}_{\text{extract}}(A, Q, \text{prompt})$
\For{each $s_i \in S$}
    \State \textbf{Compute hallucination scores:}
    \State \hspace{1.5em} $p_{ij} = M_j(s_i, Q), \quad \forall M_j$
    
    \State \textbf{Compute final probability:}
    \State \hspace{1.5em} $p_i = \frac{1}{N} \sum_{j=1}^{N} p_{ij}$

    \If{$p_i \geq \tau$}
        \State \textbf{Add hallucinated span to refined set:}
        \State \hspace{1.5em} $S' \gets S' \cup \{s_i\}$
    \EndIf
\EndFor

\State \textbf{Apply fuzzy matching for span refinement:}
\State \hspace{1em} $S^* \gets \text{FuzzyMatch}(S', \text{Ground Truth}, 0.9)$

\State \textbf{Return} $S^*$
\end{algorithmic}
\end{algorithm}


\begin{thebibliography}{22}
\providecommand{\natexlab}[1]{#1}

\bibitem[{Abdelrahman(2024)}]{abdelrahman2024hallucination}
Mostafa Abdelrahman. 2024.
\newblock Hallucination in low-resource languages: Amplified risks and
  mitigation strategies for multilingual llms.
\newblock \emph{Journal of Applied Big Data Analytics, Decision-Making, and
  Predictive Modelling Systems}, 8(12):17--24.

\bibitem[{Alaharju(2024)}]{alaharju2024ensuring}
Henri Alaharju. 2024.
\newblock Ensuring performance and reliability in llm-based applications: A
  case study.

\bibitem[{Chaudhuri et~al.(2003)Chaudhuri, Ganjam, Ganti, and
  Motwani}]{chaudhuri2003robust}
Surajit Chaudhuri, Kris Ganjam, Venkatesh Ganti, and Rajeev Motwani. 2003.
\newblock Robust and efficient fuzzy match for online data cleaning.
\newblock In \emph{Proceedings of the 2003 ACM SIGMOD international conference
  on Management of data}, pages 313--324.

\bibitem[{Hikal et~al.(2025)Hikal, Nasreldin, Hamdi, and
  Mohammed}]{hikal2025few}
Baraa Hikal, Ahmed Nasreldin, Ali Hamdi, and Ammar Mohammed. 2025.
\newblock Few-shot optimized framework for hallucination detection in
  resource-limited nlp systems.
\newblock \emph{arXiv preprint arXiv:2501.16616}.

\bibitem[{Huang et~al.(2025)Huang, Yu, Ma, Zhong, Feng, Wang, Chen, Peng, Feng,
  Qin et~al.}]{huang2025survey}
Lei Huang, Weijiang Yu, Weitao Ma, Weihong Zhong, Zhangyin Feng, Haotian Wang,
  Qianglong Chen, Weihua Peng, Xiaocheng Feng, Bing Qin, et~al. 2025.
\newblock A survey on hallucination in large language models: Principles,
  taxonomy, challenges, and open questions.
\newblock \emph{ACM Transactions on Information Systems}, 43(2):1--55.

\bibitem[{Ji et~al.(2023)Ji, Lee, Frieske, Yu, Su, Xu, Ishii, Bang, Madotto,
  and Fung}]{ji2023survey}
Ziwei Ji, Nayeon Lee, Rita Frieske, Tiezheng Yu, Dan Su, Yan Xu, Etsuko Ishii,
  Ye~Jin Bang, Andrea Madotto, and Pascale Fung. 2023.
\newblock Survey of hallucination in natural language generation.
\newblock \emph{ACM computing surveys}, 55(12):1--38.

\bibitem[{Joshi et~al.(2020)Joshi, Chen, Liu, Weld, Zettlemoyer, and
  Levy}]{joshi2020spanbert}
Mandar Joshi, Danqi Chen, Yinhan Liu, Daniel~S Weld, Luke Zettlemoyer, and Omer
  Levy. 2020.
\newblock Spanbert: Improving pre-training by representing and predicting
  spans.
\newblock \emph{Transactions of the association for computational linguistics},
  8:64--77.

\bibitem[{Kang et~al.(2024{\natexlab{a}})Kang, Blevins, and
  Zettlemoyer}]{kang2024comparing}
Haoqiang Kang, Terra Blevins, and Luke Zettlemoyer. 2024{\natexlab{a}}.
\newblock Comparing hallucination detection metrics for multilingual
  generation.
\newblock \emph{arXiv preprint arXiv:2402.10496}.

\bibitem[{Kang et~al.(2024{\natexlab{b}})Kang, Woensel, and
  Seneviratne}]{kang2024using}
Inwon Kang, William~Van Woensel, and Oshani Seneviratne. 2024{\natexlab{b}}.
\newblock Using large language models for generating smart contracts for health
  insurance from textual policies.
\newblock In \emph{AI for Health Equity and Fairness: Leveraging AI to Address
  Social Determinants of Health}, pages 129--146. Springer.

\bibitem[{Liang and et~al.(2024)}]{deepseek2024v3}
Wenfeng Liang and et~al. 2024.
\newblock Deepseek-v3 technical report.
\newblock \url{https://arxiv.org/abs/2412.19437}.
\newblock Accessed: 2025-04-26.

\bibitem[{Liu and Wang(2024)}]{liu2024genotex}
Haoyang Liu and Haohan Wang. 2024.
\newblock Genotex: A benchmark for evaluating llm-based exploration of gene
  expression data in alignment with bioinformaticians.
\newblock \emph{arXiv preprint arXiv:2406.15341}.

\bibitem[{Mehta et~al.(2024)Mehta, Hoblitzell, O’keefe, Jang, and
  Varma}]{mehta2024halu}
Rahul Mehta, Andrew Hoblitzell, Jack O’keefe, Hyeju Jang, and Vasudeva Varma.
  2024.
\newblock Halu-nlp at semeval-2024 task 6: Metacheckgpt-a multi-task
  hallucination detection using llm uncertainty and meta-models.
\newblock In \emph{Proceedings of the 18th International Workshop on Semantic
  Evaluation (SemEval-2024)}, pages 342--348.

\bibitem[{Nie(2022)}]{nie2022zero}
Ercong Nie. 2022.
\newblock Zero-shot learning on low-resource languages by cross-lingual
  retrieval.
\newblock \emph{Masterarbeit im Studiengang Computerlinguistik an der
  Ludwig-Maximilians-Universit{\"a}t Munch{\"e}n Fakult{\"a}t fur Sprach-und
  Literaturwissenschaften}.

\bibitem[{OpenAI(2024)}]{openai2024gpt4o}
OpenAI. 2024.
\newblock Gpt-4o system card.
\newblock \url{https://arxiv.org/html/2410.21276v1}.
\newblock Accessed: 2025-04-26.

\bibitem[{Rezatofighi et~al.(2019)Rezatofighi, Tsoi, Gwak, Sadeghian, Reid, and
  Savarese}]{rezatofighi2019generalized}
Hamid Rezatofighi, Nathan Tsoi, JunYoung Gwak, Amir Sadeghian, Ian Reid, and
  Silvio Savarese. 2019.
\newblock Generalized intersection over union: A metric and a loss for bounding
  box regression.
\newblock In \emph{Proceedings of the IEEE/CVF conference on computer vision
  and pattern recognition}, pages 658--666.

\bibitem[{Sheugh and Alizadeh(2015)}]{sheugh2015note}
Leily Sheugh and Sasan~H Alizadeh. 2015.
\newblock A note on pearson correlation coefficient as a metric of similarity
  in recommender system.
\newblock In \emph{2015 AI \& Robotics (IRANOPEN)}, pages 1--6. IEEE.

\bibitem[{Smith et~al.(2024)Smith, Fries, Hancock, and
  Bach}]{smith2024language}
Ryan Smith, Jason~A Fries, Braden Hancock, and Stephen~H Bach. 2024.
\newblock Language models in the loop: Incorporating prompting into weak
  supervision.
\newblock \emph{ACM/JMS Journal of Data Science}, 1(2):1--30.

\bibitem[{Sriramanan et~al.(2025)Sriramanan, Bharti, Sadasivan, Saha,
  Kattakinda, and Feizi}]{sriramanan2025llm}
Gaurang Sriramanan, Siddhant Bharti, Vinu~Sankar Sadasivan, Shoumik Saha,
  Priyatham Kattakinda, and Soheil Feizi. 2025.
\newblock Llm-check: Investigating detection of hallucinations in large
  language models.
\newblock \emph{Advances in Neural Information Processing Systems},
  37:34188--34216.

\bibitem[{Tsarfaty et~al.(2013)Tsarfaty, Seddah, K{\"u}bler, and
  Nivre}]{tsarfaty2013parsing}
Reut Tsarfaty, Djam{\'e} Seddah, Sandra K{\"u}bler, and Joakim Nivre. 2013.
\newblock Parsing morphologically rich languages: Introduction to the special
  issue.
\newblock \emph{Computational linguistics}, 39(1):15--22.

\bibitem[{V\'azquez et~al.(2025)V\'azquez, Mickus, Zosa, Vahtola, Tiedemann,
  Sinha, Segonne, S\'anchez-Vega, Raganato, Libovický, Karlgren, Ji, Helcl,
  Guillou, de~Gibert, Bengoetxea, Attieh, and
  Apidianaki}]{vazquez-etal-2025-mu-shroom}
Ra\'ul V\'azquez, Timothee Mickus, Elaine Zosa, Teemu Vahtola, J\"org
  Tiedemann, Aman Sinha, Vincent Segonne, Fernando S\'anchez-Vega, Alessandro
  Raganato, Jindřich Libovický, Jussi Karlgren, Shaoxiong Ji, Jindřich
  Helcl, Liane Guillou, Ona de~Gibert, Jaione Bengoetxea, Joseph Attieh, and
  Marianna Apidianaki. 2025.
\newblock \href {https://helsinki-nlp.github.io/shroom/} {Sem{E}val-2025 {T}ask
  3: {Mu-SHROOM}, the multilingual shared-task on hallucinations and related
  observable overgeneration mistakes}.

\bibitem[{Vectara(2024)}]{vectara2024hallucination}
Vectara. 2024.
\newblock \href
  {https://github.com/vectara/hallucination-leaderboard/commit/9708eccda25bf8640db6c6748ac25369947309ac}
  {Hallucination leaderboard}.
\newblock GitHub repository commit.

\bibitem[{Yang et~al.(2024)Yang, Yang, Zhang, Hui, Zheng, Yu, Li, Liu, Huang,
  Wei, Lin, Yang, Tu, Zhang, Yang, Yang, Zhou, Lin, Dang, Lu, Bao, Yang, Yu,
  Li, Xue, Zhang, Zhu, Men, Lin, Li, Tang, Xia, Ren, Ren, Fan, Su, Zhang, Wan,
  Liu, Cui, Zhang, and Qiu}]{qwen2024technical}
An~Yang, Baosong Yang, Beichen Zhang, Binyuan Hui, Bo~Zheng, Bowen Yu,
  Chengyuan Li, Dayiheng Liu, Fei Huang, Haoran Wei, Huan Lin, Jian Yang,
  Jianhong Tu, Jianwei Zhang, Jianxin Yang, Jiaxi Yang, Jingren Zhou, Junyang
  Lin, Kai Dang, Keming Lu, Keqin Bao, Kexin Yang, Le~Yu, Mei Li, Mingfeng Xue,
  Pei Zhang, Qin Zhu, Rui Men, Runji Lin, Tianhao Li, Tianyi Tang, Tingyu Xia,
  Xingzhang Ren, Xuancheng Ren, Yang Fan, Yang Su, Yichang Zhang, Yu~Wan,
  Yuqiong Liu, Zeyu Cui, Zhenru Zhang, and Zihan Qiu. 2024.
\newblock Qwen2.5 technical report.
\newblock \url{https://arxiv.org/abs/2412.15115}.
\newblock Accessed: 2025-04-26.

\end{thebibliography}
\end{document}